\documentclass[10pt, conference, letterpaper]{ IEEEtran}
\IEEEoverridecommandlockouts

\usepackage{amsmath,amssymb,amsthm}
\usepackage{booktabs,multirow,array}
\usepackage{graphicx}
\graphicspath{{fig/}}
\usepackage{xcolor}
\usepackage{url}
\usepackage{cite}
\usepackage{enumitem}
\usepackage{balance}
\usepackage{bm}
\usepackage{mathtools}
\usepackage{pifont}
\usepackage{algorithm}
\usepackage{subcaption}
\usepackage{algpseudocode}

\def\BibTeX{{\rm B\kern-.05em{\sc i\kern-.025em b}\kern-.08em
    T\kern-.1667em\lower.7ex\hbox{E}\kern-.125emX}}

\begin{document}

\title{TrustAgent-RFF: Agentic Diagnostic Planning for Reliable RF Fingerprinting under Domain Shift}

\IEEEoverridecommandlockouts
\makeatletter
\renewcommand{\@IEEEauthorblockconfadjspace}{-0.85em}
\makeatother

\author{
\IEEEauthorblockN{
Chuan Liu$^{1}$, Tianya Zhao$^{1}$, Yiting Wang$^{1}$,
Junqing Zhang$^{2}$, Xuyu Wang$^{1,*}$
}
\IEEEauthorblockA{\small
$^{1}$Knight Foundation School of Computing and Information Sciences,
Florida International University, Miami, FL 33199, USA\\
$^{2}$School of Computer Science and Informatics,
University of Liverpool, Liverpool, United Kingdom\\
Emails: \{cliu060, tzhao010, ywang253, xuywang\}@fiu.edu;
junqing.zhang@liverpool.ac.uk
}
}

\IEEEaftertitletext{\vspace{-6pt}}

\maketitle

\begin{abstract}
Radio frequency fingerprinting (RFF) models can achieve accuracy under source domain conditions but become unreliable when the receiver, distance, collection day, or environment changes. Incorrect predictions in unseen domains may receive high softmax confidence, making confidence thresholding insufficient for trustworthy deployment. We propose TrustAgent-RFF, a prediction level reliability framework for RFF that operates with a frozen in-phase and quadrature (I/Q) classifier without target domain labels. For each device prediction, the framework acquires radio frequency evidence: counterfactual stability, prototype consistency, acquisition domain similarity, and signal quality. A source calibrated trust estimator combines these diagnostics with classifier confidence, prediction margin, and evidence availability
indicators. Source derived lower tail penalties and cross module interactions reduce trust when diagnostic signals indicate weak support. A large language model (LLM) based planner adaptively selects the next diagnostic action based on evidence, trust state, action costs, and remaining budget. A deterministic source calibrated decision layer updates the prediction state as Accept, Reject, or Defer. Experiments across five RFF datasets and classifier backbones reveal an average source to target accuracy drop of 38.7\% and show that many incorrect target domain predictions remain highly confident.


\end{abstract}

\begin{IEEEkeywords}
Radio Frequency Fingerprinting, Agentic Diagnostic Planning, Large Language Models, Domain Shift.
\end{IEEEkeywords}

\section{Introduction}
\label{sec:introduction}

Radio frequency fingerprinting (RFF) identifies wireless transmitters
from hardware dependent distortions embedded in their emitted signals.
These distortions arise from manufacturing variations and nonideal wireless
components, such as oscillators, mixers, power amplifiers, and data
converters. Deep neural networks can learn device dependent
characteristics directly from raw in-phase and quadrature (I/Q) samples
\cite{merchant2018deep,riyaz2018deep}. More recent systems achieve high
identification accuracy when training and testing conditions are similar
\cite{sankhe2019oracle,shen2021radio_deep}.

However, the received waveform is also shaped by the receiver,
propagation channel, distance, collection day, and surrounding
environment. These factors can introduce acquisition specific artifacts
and cause distribution shifts without changing the transmitter
identity. Channel effects can greatly alter learned radio frequency (RF) representations
\cite{al2020exposing}, motivating channel resilient fingerprinting
methods
\cite{restuccia2019deepradioid}. Existing work addresses physical domain
shift through receiver domain adaptation
\cite{yang2024mitigating}, domain generalization
\cite{zhang2024domain}, and cross domain representation
learning
\cite{zhao2024cross}. These methods primarily improve the accuracy or
robustness of the underlying identity classifier.

This paper considers a complementary deployment question: \textit{after a
pretrained classifier produces a device identity, can we determine
whether that particular prediction is sufficiently reliable to be
accepted?} This question is important because classifier confidence does
not necessarily reflect correctness. Modern neural networks can be poorly calibrated
\cite{guo2017calibration}. Predictive uncertainty can degrade further
under distribution shift
\cite{ovadia2019can}, while maximum softmax probability may remain
high for misclassified or out of distribution inputs
\cite{hendrycks2016baseline}. Since a closed set classifier must assign every
sample to one of the enrolled devices, an incorrect prediction from an
unseen acquisition domain may still receive high confidence.

To address this problem, we present \textsc{TrustAgent-RFF}, a
prediction level reliability assessment framework for RFF under unseen
physical domain shifts. The framework is placed after a frozen raw IQ
classifier and does not modify its predicted identity. Instead, it
collects four complementary forms of RF evidence: counterfactual
stability, predicted device prototype consistency, acquisition domain
similarity, and signal quality. These diagnostics are fused with the
classifier confidence and prediction margin to estimate a
source calibrated trust score.

When the trust state remains unresolved, a LLM based planner selects the next affordable diagnostic using the
acquired evidence, trust score, action costs, remaining budget, and diagnostic history. This design is related to language model based tool selection \cite{schick2023toolformer}. After each action, the trust score is updated and a deterministic decision layer produces \textsc{Accept}, \textsc{Reject}, or \textsc{Defer}. The LLM does not
modify the classifier output, calculate the trust score, or determine correctness; it only controls diagnostic acquisition.

Unlike target aware adaptation methods, \textsc{TrustAgent-RFF} uses no
labeled target domain samples for classifier training, trust estimator
calibration, evidence threshold selection, or diagnostic planning. The
framework focuses on unknown domain reliability, where the transmitter
classes are present in the training label space but the target
acquisition conditions are unseen. It therefore complements existing
efforts to improve cross domain classification by providing a separate
prediction level reliability layer for deployment.

We evaluate \textsc{TrustAgent-RFF} on five RF fingerprinting datasets
covering distance, receiver, temporal, and environmental shifts, and
across multiple convolutional, recurrent, temporal, self supervised, and
Transformer based classifiers. The results show that strong source domain
performance does not guarantee target domain reliability and that many
incorrect target predictions remain highly confident. They further show
that combining multiple RF diagnostics improves the ranking and filtering
of unreliable predictions, reducing the error rate among retained
high confidence samples while preserving a useful level of coverage.
We additionally evaluate adaptive diagnostic planning against
non adaptive controller baselines to study the tradeoff between
reliability improvement and diagnostic cost.
The main contributions of this work are summarized as follows:

\begin{itemize}[leftmargin=*]

\item We formulate prediction level reliability assessment for RFF
under unseen physical domain shifts. Unlike prior work that mainly
focuses on improving average identification accuracy, our formulation
separates device prediction from deployment time trust assessment.

\item We develop four RF-specific diagnostics that capture local
decision instability, insufficient support for the predicted device,
unsimilar acquisition conditions, and degraded signal quality. These
diagnostics provide evidence beyond softmax confidence without using
labeled target domain data.

\item We develop a source calibrated trust layer and an LLM based controller. The trust layer fuses partially acquired evidence
and deterministically produces \textsc{Accept}, \textsc{Reject}, or
\textsc{Defer}, while the LLM selects only the next diagnostic action
under the available budget.

\item We evaluate \textsc{TrustAgent-RFF} across five datasets and
multiple classifier architectures, showing persistent high confidence
errors under physical domain shift and improved selective reliability
through adaptive diagnostic acquisition.

\end{itemize}
\section{Related Work}
\label{sec:related_work}

\subsection{Deep and Cross Domain RF Fingerprinting}

RFF identifies wireless transmitters from device dependent distortions
introduced by nonideal RF components. Early deep learning approaches
demonstrated that transmitter representations can be learned directly
from raw RF observations
\cite{merchant2018deep,riyaz2018deep}. Convolutional neural network (CNN) based systems have since been
developed for both raw IQ samples
\cite{sankhe2019oracle,shen2021radio_deep}
and spectrogram representations
\cite{shen2021radio_spectrogram}. DeepLoRa further combines deep
learning with signal augmentation for large scale LoRa identification
\cite{al2021deeplora}.

A major challenge is that the received waveform is affected by
the transmitter, receiver, channel, distance, collection time, and
environment. Wireless channel effects can substantially alter learned
RF fingerprints and degrade cross domain accuracy
\cite{al2020exposing}. DeepRadioID addresses related channel variation
through channel resilient radio fingerprint optimization
\cite{restuccia2019deepradioid}. Other work studies the domain
generalizability of RF representations under temporal and environmental
changes
\cite{johnson2023domain} and under backdoor attacks~\cite{zhao2024explanation, zhao2025protocol}.

Cross receiver RFF is particularly difficult because receiver hardware
introduces additional distortions. Existing methods address this problem
through domain adaptation
\cite{yang2024mitigating}, domain generalization
\cite{zhang2024domain}, and interpretable cross domain representation
learning
\cite{zhao2024cross, zhao2026generalizable}. Unlike these approaches,
\textsc{TrustAgent-RFF} does not adapt the identity classifier but
evaluates the reliability of its individual predictions.
\subsection{Reliability Assessment under Distribution Shift}

Maximum softmax probability is widely used as a confidence score, but
modern neural networks can be overconfident and miscalibrated
\cite{guo2017calibration}. Bayesian approximations estimate uncertainty
through stochastic forward passes
\cite{gal2016dropout}, while deep ensembles combine multiple predictors
to improve uncertainty estimation
\cite{lakshminarayanan2017simple}. However, uncertainty quality can still
degrade substantially under distribution shift
\cite{ovadia2019can}.

Out of distribution detection aims to distinguish atypical inputs from
those supported by the training distribution. Maximum softmax
probability provides a common baseline
\cite{hendrycks2016baseline}. ODIN improves detection through temperature
scaling and input perturbations
\cite{liang2017enhancing}, while energy based scores provide an
alternative to normalized softmax confidence
\cite{liu2020energy}. However, domain dissimilarity and prediction
correctness are not equivalent. A sample from an unseen acquisition
domain may still be classified correctly, while a familiar sample may
be misclassified near a decision boundary.

Selective classification allows a model to abstain from uncertain
predictions and evaluates the tradeoff between coverage and accepted
risk
\cite{geifman2017selective}. SelectiveNet integrates the selection
mechanism into neural network training
\cite{geifman2019selectivenet}, while one sided prediction provides
another formulation of selective decision making
\cite{gangrade2021selective}. Recent studies further investigate post
hoc confidence estimators
\cite{cattelan2023fix} and contrastive objectives for selective
classification
\cite{wu2024confidence}.

Conformal prediction uses held out calibration data to construct
statistically valid predictive procedures
\cite{angelopoulos2023conformal}. Distribution free and learn then test
frameworks further provide explicit control of selected risk criteria
\cite{bates2021distribution,angelopoulos2025learn}.
\textsc{TrustAgent-RFF} adopts the selective prediction perspective but
uses RFF specific diagnostic evidence to assess individual predictions
under physical domain shift.
\section{Motivation}
\label{sec:background_motivation}

RF fingerprints are affected by propagation channels, receiver
hardware, and acquisition conditions. Channel effects can alter learned
transmitter representations
\cite{al2020exposing}, while receiver dependent distortions reduce cross
receiver generalization
\cite{yang2024mitigating}. Changes in distance, receiver, collection
day, or environment can therefore shift the signal distribution and
degrade source trained models. Recent domain generalization studies
confirm the difficulty of transferring RFF models to unseen physical
conditions
\cite{zhang2024domain}. Our experiments on Bluetooth, ORACLE, WiSig,
LoRa, and NetSTAR LoRa in Section~V show substantial but dataset
dependent cross domain degradation, indicating that source accuracy
alone is insufficient to characterize deployment reliability.

\begin{figure}[t]
    \centering

    \begin{subfigure}[t]{0.39\columnwidth}
        \centering
        \includegraphics[
            height=3.0cm
        ]{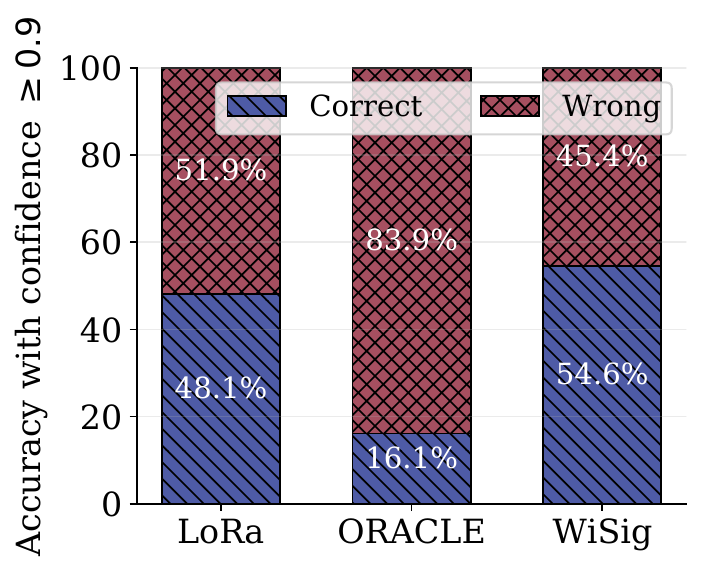}
        \caption{Composition of high confidence predictions.}
        \label{fig:motivation_high_confidence}
    \end{subfigure}
    \hfill
    \begin{subfigure}[t]{0.59\columnwidth}
        \centering
        \includegraphics[
            height=3.0cm,
            trim=7 0 10 0,
            clip
        ]{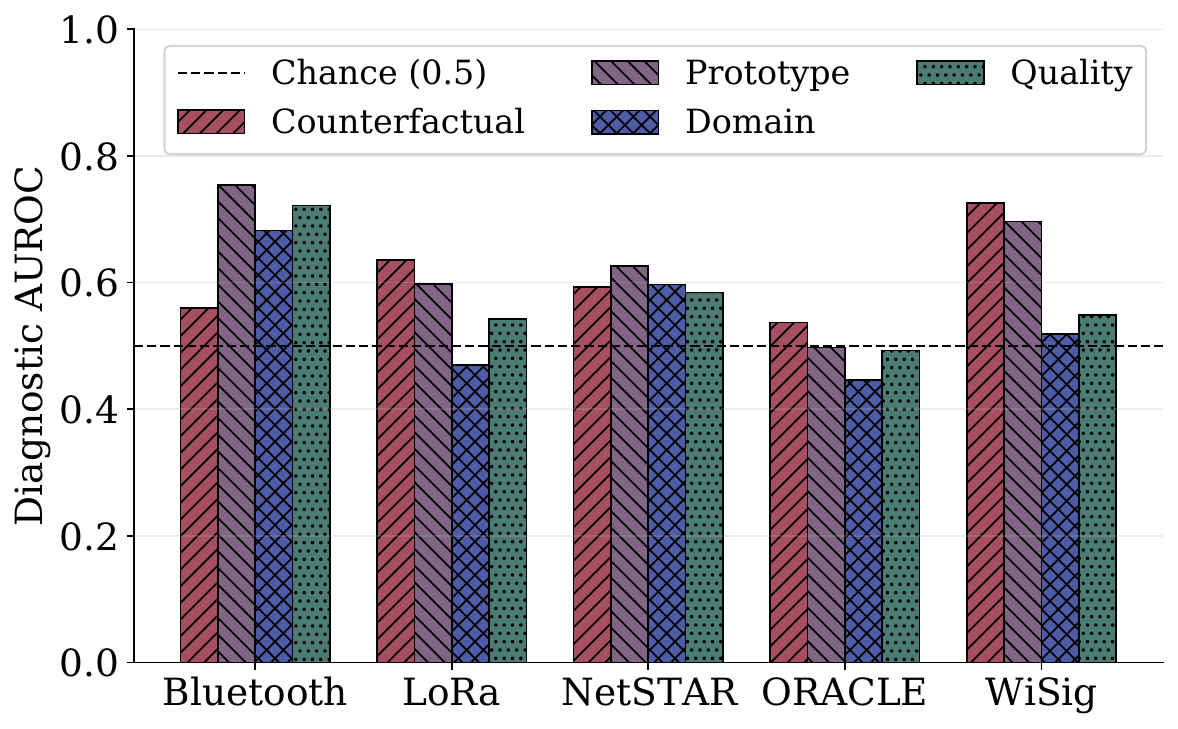}
        \caption{Performance of individual diagnostics.}
        \label{fig:motivation_diagnostic}
    \end{subfigure}

    \caption{
Motivation for reliability aware RFF inference.
(a) High confidence predictions can remain incorrect under unseen
domain shifts.
(b) Diagnostics provide complementary reliability evidence.
}
    \label{fig:motivation}
\end{figure}

Maximum softmax can remain high for misclassified or out of
distribution inputs
\cite{guo2017calibration,ovadia2019can,hendrycks2016baseline}.
Softmax confidence reflects only the relative preference among enrolled
devices and does not indicate whether a sample is supported by the
source distribution. As shown in
Fig.~\ref{fig:motivation_high_confidence}, among target predictions with
confidence $\geq 0.9$, the error rates remain 51.9\% on LoRa, 83.9\% on
ORACLE, and 45.4\% on WiSig. Thus, confidence thresholding alone cannot
reliably identify incorrect predictions and often reduces errors mainly
by lowering coverage.

These observations motivate RFF specific diagnostic evidence.
Figure~\ref{fig:motivation_diagnostic} shows that counterfactual
stability, prototype consistency, domain similarity, and signal quality
vary in error detection performance across datasets, with no diagnostic
consistently dominating. \textsc{TrustAgent-RFF} therefore combines
them as complementary evidence.

Computing all diagnostics for every prediction is costly.
\textsc{TrustAgent-RFF} instead uses an LLM planner to select the next
affordable diagnostic when the current trust state remains unresolved.
This follows language model based tool selection, where an LLM selects
external operations rather than producing the final task output
\cite{schick2023toolformer}. Different predictions can therefore follow
different diagnostic trajectories according to their evidence and
remaining budget.

Finally, forcing a device identity for every input can be unsafe in
authentication and security applications. Selective classification
allows abstention and evaluates the tradeoff between coverage and risk
\cite{geifman2017selective}, while conformal and risk controlling
methods use calibration data to support statistical risk criteria
\cite{angelopoulos2023conformal,bates2021distribution}.
Following this perspective, \textsc{TrustAgent-RFF} outputs
\textsc{Accept}, \textsc{Reject}, or \textsc{Defer} rather than
accepting every prediction.

\section{TrustAgent-RFF}
\label{sec:method}

\subsection{Overview and Problem Formulation}
\label{subsec:overview}

Let $x\in\mathbb{R}^{2\times L}$ denote a raw IQ window and
$f_\theta$ a pretrained RF fingerprinting classifier. The classifier
produces
$
f_\theta(x)=(\hat y,\bm p,\bm z),
$
where $\hat y=\arg\max_k p_k$ is the predicted transmitter identity,
$\bm p$ is the class posterior vector, and
$\bm z$ is an intermediate representation.
 We define
the confidence and prediction margin as
$
c=\max_k p_k$,
$m=p_{(1)}-p_{(2)}$, where $p_{(1)}$ and $p_{(2)}$ are the two largest entries of $\bm p$.

\begin{figure}[t]
\centering
\includegraphics[
  width=\columnwidth,
  trim={25mm 35mm 18mm 35mm},
  clip
]{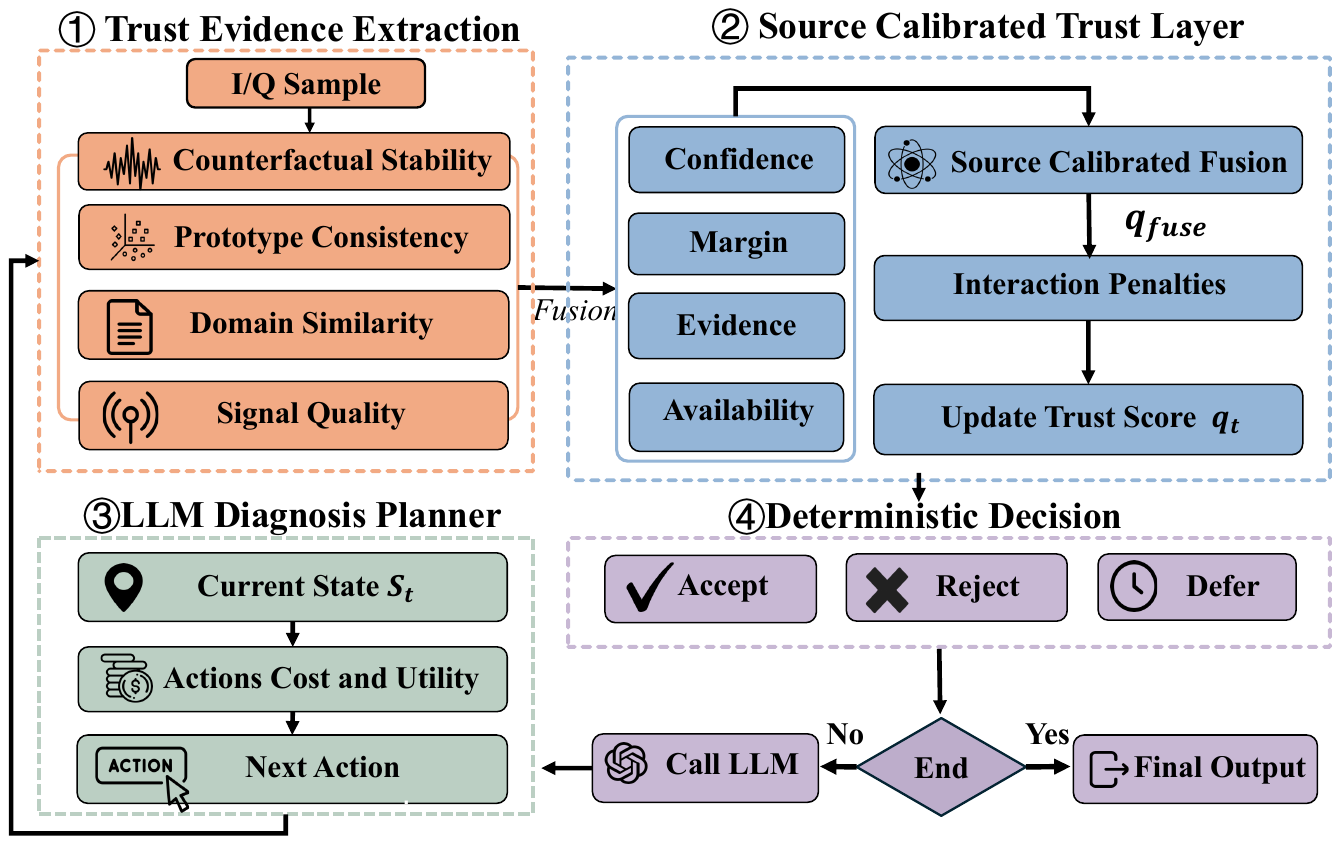}
\caption{
Overview of \textsc{TrustAgent-RFF}. The framework acquires
complementary IQ based diagnostic evidence, updates a source calibrated
trust score, and uses an LLM planner to select diagnostics
when the evidence is insufficient. Deterministic decision
layer outputs \textsc{Accept}, \textsc{Reject}, or \textsc{Defer}.
}
\label{fig:overview}
\end{figure}

As shown in Fig.~\ref{fig:overview}, \textsc{TrustAgent-RFF} operates
after the frozen classifier and does not modify $\hat y$. Its output is
\begin{equation}
x
\xrightarrow{f_\theta}
(\hat y,\bm p,\bm z)
\xrightarrow{\textsc{TrustAgent-RFF}}
(\hat y,q,d,\bm e,\bm h),
\label{eq:framework}
\end{equation}
where $q$ is the trust score,
$d\in\{\textsc{Accept},\textsc{Reject},\textsc{Defer}\}$ is the
deployment decision, $\bm e$ contains the acquired diagnostic evidence,
and $\bm h$ records the diagnostic trajectory. The framework provides four diagnostic actions:
\begin{equation}
\mathcal A_{\rm diag}
=
\left\{
A_{\rm cf},
A_{\rm proto},
A_{\rm dom},
A_{\rm qual}
\right\},
\end{equation}
corresponding to counterfactual stability, prototype consistency,
domain similarity, and signal quality. These modules respectively
measure local prediction stability, support for the predicted device,
similarity to source acquisition domains, and consistency with source
IQ statistics. No individual diagnostic is interpreted as a direct
correctness detector.

The acquired evidence is combined with classifier confidence and margin
to update the trust score. A deterministic decision layer produces a
preliminary \textsc{Accept}, \textsc{Reject}, or \textsc{Defer} state.
A terminal \textsc{Accept} or \textsc{Reject} is returned only when the
required diagnostic condition is satisfied. Otherwise, an LLM planner
selects the next affordable diagnostic or stops the trajectory. The
planner controls only diagnostic acquisition and does not modify the
classifier output, calculate the trust score, or directly determine
prediction correctness.

For labeled source calibration samples, prediction correctness is
determined by whether $\hat y(x)=y$. During target deployment, the
ground-truth label is unavailable.
\textsc{TrustAgent-RFF} therefore estimates reliability using the frozen
classifier, source calibrated statistics, and evidence extracted from
the current IQ sample. The trust estimator, decision thresholds, and
diagnostic controller are configured without labeled target domain
samples.

\subsection{Trust Evidence Extraction}
\subsubsection{Counterfactual Stability}
\label{subsubsec:counterfactual}

The counterfactual action evaluates the local stability of the
classifier under bounded perturbations of the observed IQ signal. Each
trial applies one or more numerical transformations, such as amplitude
scaling, phase rotation, additive noise, time displacement, or a phase
ramp. These transformations probe the sensitivity of the classifier
around the current input and are not intended to estimate specific
physical impairments.

Let $\widetilde{x}_j=T_j(x)$ denote the sample generated by the $j$th
RF perturbation, where $j\in\{1,\ldots,M\}$. We use $M=20$
counterfactual perturbations for each input sample. Let
$
\hat y(x)=\arg\max_k p_k(x)
$
denote the original prediction. The counterfactual stability score is
\begin{equation}
s_{\rm cf}(x)
=
\frac{1}{M}
\sum_{j=1}^{M}
\mathbb{I}\!\left[
\arg\max_k p_k(\widetilde{x}_j)
=
\hat y(x)
\right].
\label{eq:counterfactual_support}
\end{equation}

A large $s_{\rm cf}(x)$ indicates local decision invariance but does not
guarantee correctness, since a domain shifted sample may remain inside
a stable but incorrect decision region. Counterfactual stability is
therefore interpreted jointly with the remaining diagnostic evidence.

\subsubsection{Prototype Consistency}
\label{subsubsec:prototype}

The prototype consistency action evaluates whether the representation
of a test sample is supported by source samples from the device
predicted by the classifier. It uses the intermediate representation of
the frozen classifier and does not require target domain labels.

Let
$
\bar{\bm z}(x)
$
denote the normalized classifier embedding. Let
$\mathcal I_k^{\rm src}$ denote the indices of source training samples
belonging to device $k$. The corresponding source prototype is
\begin{equation}
\widetilde{\bm\mu}_k
=
\frac{1}{|\mathcal I_k^{\rm src}|}
\sum_{i\in\mathcal I_k^{\rm src}}
\bar{\bm z}(x_i),
\qquad
\bm\mu_k
=
\frac{\widetilde{\bm\mu}_k}
{\lVert\widetilde{\bm\mu}_k\rVert_2}.
\label{eq:prototype_center}
\end{equation}

For a test sample whose classifier prediction is $\hat y$, the
prototype distance is
$
d_{\rm proto}(x)
=
\left\|
\bar{\bm z}(x)-\bm\mu_{\hat y}
\right\|_2.
$
The predicted device prototype is used as the reference rather than the
ground truth target identity. Therefore, the diagnostic evaluates the
representation level support for the classifier output available at
deployment time.

For each source device $k$, an empirical reference distribution is
constructed from the distances
$\left\{
\left\|
\bar{\bm z}(x_i)-\bm\mu_k
\right\|_2
:
i\in\mathcal I_k^{\rm src}
\right\}.
$ Let $\widehat F_k^{\rm proto}$ denote the empirical cumulative
distribution function of these source distances. Prototype
consistency score is
\begin{equation}
s_{\rm proto}(x)
=
1-
\widehat F_{\hat y}^{\rm proto}
\left(
d_{\rm proto}(x)
\right).
\label{eq:prototype_model}
\end{equation}

A large $s_{\rm proto}(x)$ indicates that the test embedding is typical
of source samples from the predicted device, whereas a small score
indicates weak representation level support for the classifier output.
Prototype consistency does not directly determine prediction
correctness and is therefore interpreted jointly with the remaining
diagnostic evidence.

\subsubsection{Domain Similarity}
\label{subsubsec:domain}

The domain similarity action evaluates whether the representation of a
test signal is supported by acquisition conditions observed in the
source data. Source domain labels may correspond to collection day,
receiver, distance, or scene, depending on the dataset. The score
measures similarity to the source acquisition domains and does not
identify the physical cause of a domain shift.

The encoder is trained with domain classification, reconstruction, and
supervised contrastive objectives, together with a device adversarial
head:
\begin{equation}
\mathcal L
=
\mathcal L_{\rm dom}
+
\lambda_{\rm rec}\mathcal L_{\rm rec}
+
\lambda_{\rm con}\mathcal L_{\rm supcon}
+
\mathcal L_{\rm dev}.
\end{equation}
A gradient reversal layer with strength $\lambda_{\rm adv}$ is placed
before the device head. It allows the device head to predict device
identity while reversing the corresponding gradient to the encoder,
thereby encouraging the learned representation to preserve acquisition
domain information while suppressing device specific information
\cite{ganin2016domain}. We set
$\lambda_{\rm rec}=0.05$, $\lambda_{\rm con}=0.30$, and
$\lambda_{\rm adv}=0.10$, with a supervised contrastive temperature of
$0.10$. Let $e_\phi(x)$ denote the representation produced by the trained
domain encoder.

For each source domain $d\in\mathcal D_{\rm src}$, we estimate a robust
center $\bm\nu_d$ and regularized diagonal variance
$\widetilde{\bm\gamma}_d^2$. The distance to the nearest source domain is
\begin{equation}
d_{\rm dom}(x)
=
\min_{d\in\mathcal D_{\rm src}}
\left[
\sum_{\ell}
\frac{
\left(e_{\phi,\ell}(x)-\nu_{d,\ell}\right)^2
}{
\widetilde{\gamma}_{d,\ell}^{\,2}
}
\right]^{1/2}.
\end{equation}

Following the same source calibrated distance to score transformation
used for prototype consistency, we define
\begin{equation}
s_{\rm dom}(x)
=
1-
\widehat F_{\rm cal}^{\rm dom}
\left(
d_{\rm dom}(x)
\right),
\label{eq:domain_model}
\end{equation}
where $\widehat F_{\rm cal}^{\rm dom}$ is the empirical CDF of nearest
source domain distances on the source calibration set.

A large $s_{\rm dom}(x)$ indicates stronger support from at least one
known acquisition domain. Low domain similarity is treated as contextual
reliability evidence rather than a direct indicator of
misclassification.
\subsubsection{Signal Quality}
\label{subsubsec:quality}

The signal quality action evaluates whether the observed IQ waveform
exhibits statistical characteristics supported by source domain data.
Unlike the prototype and domain modules, it operates directly on the
raw IQ samples and does not rely on a learned representation.

For an IQ window $x=[I,Q]\in\mathbb{R}^{2\times L}$, we construct a
feature vector $\bm v(x)\in\mathbb{R}^{J}$ from raw signal statistics
describing signal power, amplitude variation, I/Q balance and
correlation, temporal variation, and invalid waveform indicators. These
features measure statistical consistency with source domain signals and
are not interpreted as estimates of specific physical impairments.

Source samples are divided into fitting and calibration subsets. The
fitting subset is used to estimate a robust location
$\widetilde{\bm\mu}_{\rm qual}$ and scale
$\widetilde{\bm\sigma}_{\rm qual}$ for each feature. For a test sample,
the normalized quality distance is
\begin{equation}
d_{\rm qual}(x)
=
\left[
\frac{1}{J}
\sum_{j=1}^{J}
\left(
\frac{
v_j(x)-\widetilde{\mu}_{{\rm qual},j}
}{
\widetilde{\sigma}_{{\rm qual},j}
}
\right)^2
\right]^{1/2}.
\end{equation}

The signal quality score is defined as
\begin{equation}
s_{\rm qual}(x)
=
1-
\widehat F_{\rm cal}^{\rm qual}
\left(
d_{\rm qual}(x)
\right),
\label{eq:quality_model}
\end{equation}
where $\widehat F_{\rm cal}^{\rm qual}$ is the empirical CDF of quality
distances computed on the source calibration set.

A large $s_{\rm qual}(x)$ indicates that the waveform is statistically
consistent with source supported IQ characteristics, whereas a small
value indicates an atypical signal. The module may report the feature
group with the largest normalized deviation as a statistical
attribution. This attribution identifies which observed statistics
deviate most from the source reference and should not be interpreted as
a verified physical cause.

\subsection{Source Calibrated Trust Layer}

The trust layer combines classifier confidence with partially acquired
diagnostic evidence to estimate prediction reliability. It consists of
a source calibrated trust estimator and evidence penalties for
atypical or conflicting diagnostic signals.

\subsubsection{Source Calibrated Trust Estimation}
\label{subsec:trust_estimation}

TrustAgent-RFF combines classifier information and the currently
available RF diagnostics using a source calibrated trust estimator.
The four diagnostic scores are
$s_{\rm cf}$, $s_{\rm proto}$, $s_{\rm dom}$, and $s_{\rm qual}$.
Because diagnostics are acquired sequentially, not every score is
available at every decision step.

For diagnostic $j$, an availability indicator $a_{j,t}$ equals one when
$s_j$ has been acquired by step $t$ and zero otherwise. When evidence is
unavailable, its source median is used as a neutral placeholder, while
$a_{j,t}$ informs the estimator that the value was not observed.

At diagnostic state $\bm a_t$, the trust feature vector is
$\bm\phi(x,\bm a_t)
=
\left[
c,
m,
\widetilde{\bm s}_t^\top,
\bm a_t^\top
\right]^\top,
$where $c=\max_k p_k$ is the original classifier confidence, $m$ is the
top two posterior margin, and $\widetilde{\bm s}_t$ contains either the
acquired diagnostic score or its source median.

An $\ell_2$-regularized logistic estimator is trained using only
source calibration predictions:

\begin{equation}
q_{\rm fuse}(x,\bm a_t)
=
\sigma
\left(
b+
\bm w^\top
\frac{
\bm\phi(x,\bm a_t)-\bm\mu_\phi
}{
\bm\sigma_\phi
}
\right).
\label{eq:source_trust_estimator}
\end{equation}

Here, $\bm\mu_\phi$ and $\bm\sigma_\phi$ denote the feature-wise mean
and standard deviation estimated from source calibration data, while
$\bm w$ and $b$ are the learned logistic regression parameters.

To support adaptive diagnostic planning, the estimator is trained on
source calibration examples with different evidence-availability
patterns. For each pattern, unavailable diagnostic values are replaced
by their source medians, and the corresponding availability indicators
are set to zero. This allows the estimator to evaluate partially
observed evidence states produced by different diagnostic trajectories.

Before any diagnostic is acquired, the controller initializes the current
trust state as $q_0=c$. Once diagnostic evidence becomes available,
$q_{\rm fuse}(x,\bm a_t)$ is computed using the source-calibrated trust
estimator. As evidence becomes available, the
score is updated by the source calibrated trust estimator using the
classifier confidence, prediction margin, acquired diagnostic evidence,
and availability mask.

\subsubsection{Source Calibrated Evidence Penalties}
\label{subsec:evidence_penalties}

The logistic estimator models the ordinary relationship between
classifier outputs, acquired evidence, and source correctness.
\textsc{TrustAgent-RFF} additionally applies one sided penalties when
available diagnostic evidence falls substantially below that observed
for correct source predictions.

For each diagnostic $j$, the $5$th, $10$th, and $20$th percentiles of
correct source calibration scores define a monotonic piecewise severity
mapping $\rho_j(x)\in[0,1]$. Severity is zero above the $20$th percentile,
increases from $0$ to $0.25$ between the $20$th and $10$th percentiles,
from $0.25$ to $0.60$ between the $10$th and $5$th percentiles, and
increases toward $1$ below the $5$th percentile. When fewer than 20
correct source calibration predictions are available, all source
calibration predictions are used as the reference.

Only sufficiently severe deviations contribute to the individual
penalty:
\begin{equation}
P_{\rm sev}(x,\bm a_t)
=
\eta_{\rm sev}
\sum_{j=1}^{4}
a_{j,t}\lambda_j
\frac{[\rho_j(x)-\tau_{\rm sev}]_+}
{1-\tau_{\rm sev}}
\label{eq:severe_evidence_penalty}
\end{equation}
where $a_{j,t}$ indicates evidence availability, $\lambda_j$ denotes
the corresponding diagnostic weight in
$\{\lambda_{\rm cf},\lambda_{\rm proto},
\lambda_{\rm dom},\lambda_{\rm qual}\}$,
$\tau_{\rm sev}$ is the severity activation threshold, and
$\eta_{\rm sev}$ controls the overall penalty scale.

Domain dissimilarity is treated as contextual evidence and is penalized
only when the available prototype or counterfactual evidence is also
inconsistent. Let
\begin{equation}
\bar r_t
=
\frac{
a_{{\rm proto},t}(1-s_{\rm proto})
+
a_{{\rm cf},t}(1-s_{\rm cf})
}{
a_{{\rm proto},t}+a_{{\rm cf},t}
},
\label{eq:available_inconsistency}
\end{equation}
with $\bar r_t=0$ when neither diagnostic is available. The combined
interaction penalty is
\begin{equation}
\begin{aligned}
P_{\rm int}(x,\bm a_t)
={}&
a_{{\rm dom},t}\lambda_{\rm int}^{\rm dom}
(1-s_{\rm dom})\bar r_t
\\
&+
a_{{\rm qual},t}a_{{\rm cf},t}
\lambda_{\rm int}^{\rm qual,cf}
\\
&\times
(1-s_{\rm qual})(1-s_{\rm cf}).
\end{aligned}
\label{eq:evidence_interaction_penalties}
\end{equation}

The final trust score is
\begin{equation}
\begin{aligned}
q(x,\bm a_t)
=
\operatorname{clip}_{[0,1]}
\Bigl(
& q_{\rm fuse}(x,\bm a_t)
-
P_{\rm sev}(x,\bm a_t)
\\
& -
P_{\rm int}(x,\bm a_t)
\Bigr).
\end{aligned}
\label{eq:final_trust_score}
\end{equation}

The severity weights are fixed nonnegative hyperparameters. The interaction weights are selected from $\{0,0.06,0.12\}$ for $\lambda_{\rm int}^{\rm dom}$ and $\{0,0.05,0.10\}$ for $\lambda_{\rm int}^{\rm qual,cf}$ using source calibration data only. The selected interaction is retained only when it reduces
source calibration AURC by at least $0.002$ without reducing selective
accuracy at the specified coverage; otherwise, both interaction weights
are set to zero.

Given source calibrated thresholds
$\tau_{\rm rej}<\tau_{\rm acc}$, predictions with
$q\geq\tau_{\rm acc}$ are accepted, those with
$q\leq\tau_{\rm rej}$ are rejected, and the remaining predictions are
deferred. The thresholds are selected using source calibration data,
with $\tau_{\rm acc}$ constrained by the prescribed minimum coverage.

\subsection{LLM Diagnosis Planner}
\label{subsec:controller}

The controller consists of an LLM-based planner and a
deterministic decision mechanism. The LLM selects diagnostic actions
and controls stopping, while trust estimation and the final
\textsc{Accept}, \textsc{Reject}, or \textsc{Defer} decision remain
deterministic. If the deterministic decision condition is not satisfied, control is
returned to the LLM planner to select the next diagnostic action. To avoid computing all diagnostics for every prediction,
\textsc{TrustAgent-RFF} performs diagnostic selection under action
count and cost budgets. At step $t$, the planner observes
\begin{equation}
\mathcal S_t
=
\left(
\hat y,c,m,q_t,
\bm e_t,\bm a_t,
N_t,C_t,\bm h_t
\right),
\label{eq:controller_state}
\end{equation}
where $\hat y$, $c$, and $m$ denote the classifier prediction,
confidence, and margin; $q_t$ is the current trust score; $\bm e_t$ and
$\bm a_t$ contain the acquired evidence and availability mask; $N_t$
and $C_t$ are the remaining action count and cost budgets; and
$\bm h_t$ records the action history. Unacquired evidence is hidden from the planner, which also receives the
decision thresholds and action costs. The affordable unexecuted actions are
\begin{equation}
\mathcal U_t
=
\left\{
A_j\in\mathcal A_{\rm diag}
\;\middle|\;
a_{j,t}=0,\;
N_t>0,\;
\kappa(A_j)\leq C_t
\right\},
\label{eq:available_actions}
\end{equation}
where $\kappa(A_j)$ is the assigned ordinal cost of diagnostic action
$A_j$. Prototype consistency and signal quality have unit cost, domain
similarity has cost 2, and counterfactual stability has cost 4.
These costs approximate relative computational demand for budget aware
planning and do not represent measured wall clock latency.

To prevent premature termination on high confidence predictions, the
minimum evidence requirement is
\begin{equation}
N_{\min}(c)
=
\begin{cases}
N_{\rm hc}, & c\geq\tau_{\rm hc},\\
N_{\rm base}, & c<\tau_{\rm hc},
\end{cases}
\label{eq:minimum_diagnostics}
\end{equation}
where $\tau_{\rm hc}=0.8$ is the controller verification threshold.
We set $N_{\rm hc}=4$ and $N_{\rm base}=1$, requiring high confidence
predictions to acquire all four diagnostics before termination, while
lower-confidence predictions require at least one diagnostic.

The stop action becomes available only when
$\sum_{j=1}^{4} a_{j,t}\geq N_{\min}(c)$. When the planner selects
$A_t\in\mathcal U_t$, the corresponding diagnostic is revealed, the
budgets are updated as $N_{t+1}=N_t-1$ and
$C_{t+1}=C_t-\kappa(A_t)$, and the trust score is recomputed using
Eq.~\eqref{eq:final_trust_score}. The procedure returns
\textsc{Accept} or \textsc{Reject} once the evidence
requirement is satisfied, and \textsc{Defer} when planner stops, budget is large, or no action remains affordable.

The LLM receives neither raw IQ samples nor target labels, prediction
correctness, or unacquired evidence. Diagnostic outputs are precomputed
and revealed only when selected, enabling controlled comparison among
planning policies. Algorithm~\ref{alg:sequential_controller} first initializes the trust score and
budgets. At each iteration, it checks whether the current state supports
a terminal decision; otherwise, it identifies affordable unexecuted
diagnostics and asks the LLM planner to select the next action or stop.
After each selected action, the corresponding evidence is revealed, the
budgets are updated, and the trust score is recomputed. The controller
returns \textsc{Defer} when the planner stops or no diagnostic remains
affordable.

\begin{algorithm}[t]

\caption{Diagnostic Control in \textsc{TrustAgent-RFF}}
\label{alg:sequential_controller}
\begin{algorithmic}[1]
\Require $x$, $(\hat y,c,m)$, budgets $(N_{\max},C_{\max})$
\Ensure $d\in\{\textsc{Accept},\textsc{Reject},\textsc{Defer}\}$

\State $\bm e,\bm a\gets\bm 0$,
$\bm h\gets\varnothing$
\State $q\gets c$,
$(N,C)\gets(N_{\max},C_{\max})$

\While{\textbf{true}}
    \State $d\gets\textsc{Decision}(q)$
    \If{$d\neq\textsc{Defer}$ \textbf{and}
    $\sum_{j=1}^{4}a_j\geq N_{\min}(c)$}
        \State \Return $d$
    \EndIf

    \State $\mathcal U\gets
    \{A_j\in\mathcal A_{\rm diag}
    \mid a_j=0,\ N>0,\ \kappa(A_j)\leq C\}$
    \If{$\mathcal U=\varnothing$}
        \State \Return \textsc{Defer}
    \EndIf

    \State $\mathcal A\gets\mathcal U$
    \If{$\sum_{j=1}^{4}a_j\geq N_{\min}(c)$}
        \State $\mathcal A\gets\mathcal A\cup\{A_{\rm stop}\}$
    \EndIf

    \State $\mathcal S\gets
    (\hat y,c,m,q,\bm e,\bm a,N,C,\bm h)$
    \State $A_t\gets\textsc{LLMPlanner}(\mathcal S,\mathcal A)$
    \If{$A_t=A_{\rm stop}$}
        \State \Return \textsc{Defer}
    \EndIf
    
    \State $e_{A_t}\gets\textsc{AcquireDiagnostic}(x,A_t)$
    \State Update $(\bm e,\bm a,\bm h)$ using $(A_t,e_{A_t})$
    \State $N\gets N-1$, $C\gets C-\kappa(A_t)$
    \State $q\gets\textsc{UpdateTrust}(c,m,\bm e,\bm a)$
\EndWhile
\end{algorithmic}
\end{algorithm}

\section{Experimental Evaluation}\label{sec:ex_eva}

\subsection{Experiment Setup}
\begin{figure}[t]
\centering

\begin{minipage}[t]{0.64\columnwidth}
\vspace{0pt}
\centering

\captionof{table}{
Dataset properties and physical domain evaluation protocols.
}
\label{tab:dataset_protocol}

\scriptsize
\setlength{\tabcolsep}{2.2pt}
\renewcommand{\arraystretch}{1.18}

\resizebox{\linewidth}{!}{
\begin{tabular}{
l
c
c
c
c
}
\toprule
\textbf{Dataset} & \textbf{Dev.} & \textbf{$N$}
& \textbf{Src.} & \textbf{Tgt.} \\
\midrule

ORACLE~\cite{sankhe2019oracle}
& 12 & 256K
& 8/20/32/50
& 14/26/38/44 \\

NetSTAR~\cite{elmaghbub2021lora}
& 20 & 240K
& D1--3
& D4--5 \\

WiSig~\cite{hanna2022wisig}
& 120 & 511K
& D1--2, RX0--8
& D3--4, RX0--8 \\

LoRa~\cite{shen2021radio_deep}
& 25 & 68K
& D1/3/4
& D2/5 \\

Bluetooth
& 18 & 52K
& S0
& S1--19 \\

\bottomrule
\end{tabular}
}

\vspace{0.6mm}
\scriptsize

\end{minipage}
\hfill
\begin{minipage}[t]{0.32\columnwidth}
\vspace{0pt}
\centering

\includegraphics[
    width=\linewidth
]{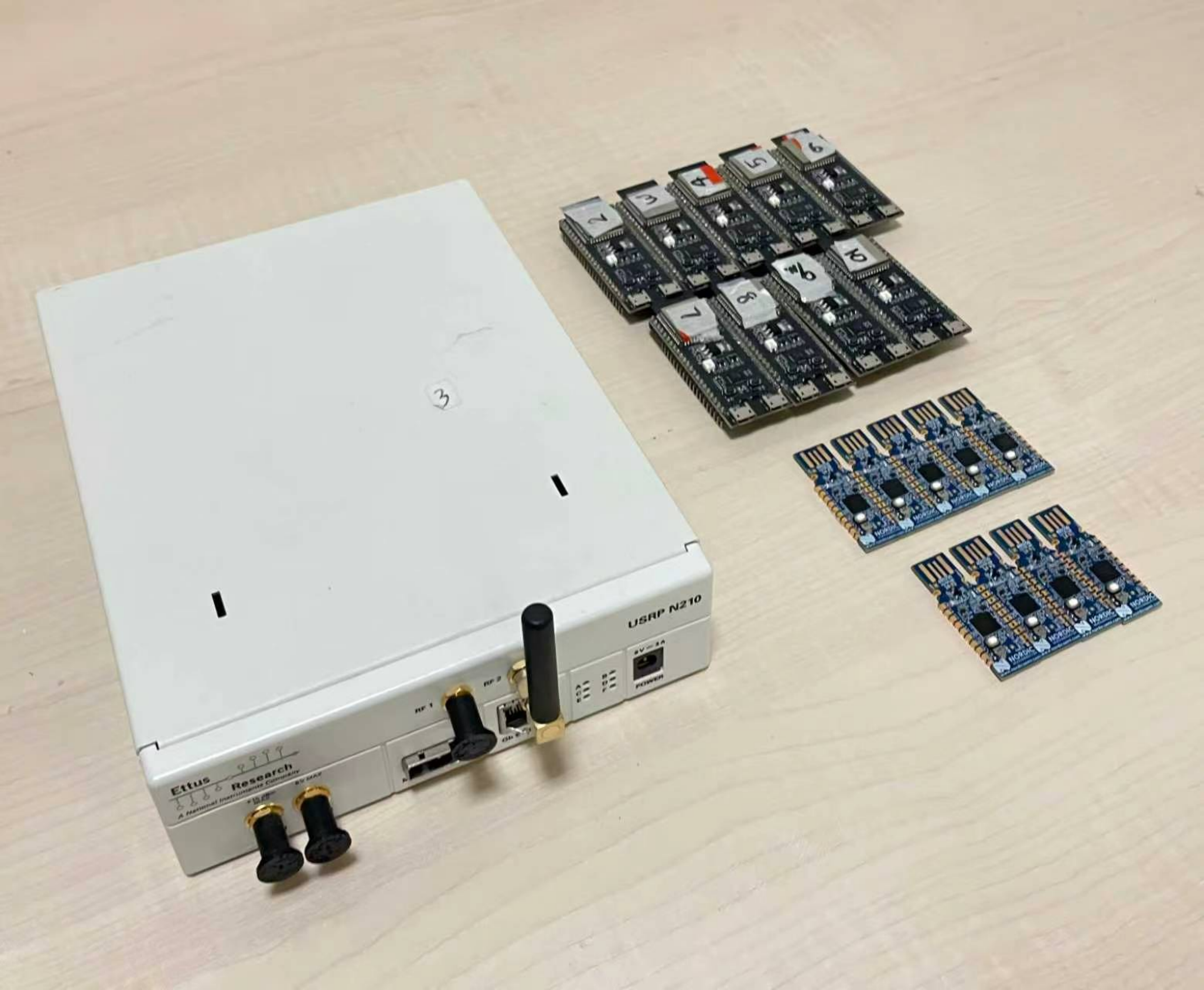}

\captionof{figure}{
BLE Dataset Devices.
}
\label{fig:dataset_devices}

\end{minipage}

\end{figure}

We evaluate \textsc{TrustAgent-RFF} on five RF fingerprinting datasets
covering distance, collection day, receiver, and acquisition scene
shifts. Table~\ref{tab:dataset_protocol} summarizes the dataset
properties and source target protocols.

\begin{table*}[t]
\centering
\caption{
Classification accuracy under source and unseen target domains.
The best result in each column is highlighted in bold. All backbones are
evaluated under our unified raw IQ training and evaluation protocol.
}
\label{tab:source_target_accuracy}
\renewcommand{\arraystretch}{1.08}
\small
\resizebox{\textwidth}{!}{
\begin{tabular}{l|cc|cc|cc|cc|cc}
\specialrule{1.1pt}{0pt}{0pt}
\textbf{Dataset $\rightarrow$}
& \multicolumn{2}{c|}{\textbf{ORACLE}}
& \multicolumn{2}{c|}{\textbf{WiSig}}
& \multicolumn{2}{c|}{\textbf{LoRa}}
& \multicolumn{2}{c|}{\textbf{NetSTAR}}
& \multicolumn{2}{c}{\textbf{Bluetooth}} \\
\hline
\textbf{Method $\downarrow$}
& Source & Target
& Source & Target
& Source & Target
& Source & Target
& Source & Target \\
\hline

CNNGRU~\cite{cho2014learning}
& 0.8575 & 0.0924
& 0.8970 & 0.4753
& 0.5290 & 0.2159
& 0.4340 & 0.4355
& 0.8417 & 0.4303 \\

PatchLLM~\cite{zhou2023one}
& 0.8717 & \textbf{0.2540}
& 0.8861 & 0.4967
& 0.5360 & 0.0392
& 0.4231 & 0.4102
& 0.9714 & 0.6777 \\

ResNet1D~\cite{he2016deep}
& 0.8638 & 0.1138
& 0.8868 & 0.3903
& 0.4225 & 0.1520
& 0.4493 & 0.4367
& 0.9858 & 0.6958 \\

SimCLR~\cite{chen2020simple}
& 0.8393 & 0.1081
& \textbf{0.9109} & 0.3317
& 0.5634 & \textbf{0.3197}
& 0.5352 & 0.4561
& 0.9681 & \textbf{0.7066} \\

TCN~\cite{bai2018empirical}
& 0.9053 & 0.0960
& 0.9059 & \textbf{0.5058}
& 0.4570 & 0.1306
& \textbf{0.6511} & 0.3832
& 0.9717 & 0.6646 \\

Transformer~\cite{vaswani2017attention}
& \textbf{0.9265} & 0.0952
& 0.8830 & 0.4889
& \textbf{0.6072} & 0.2347
& 0.4659 & \textbf{0.5269}
& \textbf{0.9900} & 0.4758 \\


\specialrule{1.1pt}{0pt}{0pt}
\end{tabular}
}
\end{table*}

ORACLE evaluates cross distance generalization, while NetSTAR and LoRa
use cross day protocols. WiSig combines cross day and receiver
variation, with one model trained on Days 1--2 over RX0--8 and tested
on Days 3--4 over the same receiver set. We collect a BLE dataset comprising signals from nine Nordic Semiconductor nRF52840 Dongles and nine Espressif Systems ESP32-S3 devices at different locations. One scene serves as the source domain, while the remaining 19 scenes serve as unseen target domains. Device identities are shared
between source and target domains in all experiments. 

To standardize the model input, all signals are segmented into fixed $2\times256$ raw IQ
windows. Same domain experiments
adopt a class stratified 70\%/10\%/20\% train, validation, and test
split. Cross domain models are trained and selected using source data
only and evaluated directly on unseen target domains. No labeled target
samples are used for training, validation, trust calibration, or
threshold selection in the primary zero shot setting.

We evaluate CNNGRU~\cite{cho2014learning},
ResNet1D~\cite{he2016deep},
TCN~\cite{bai2018empirical},
Transformer~\cite{vaswani2017attention},
SimCLR~\cite{chen2020simple}, and
PatchLLM~\cite{zhou2023one}. Models are trained for up to
50 epochs using AdamW with batch size 512, learning rate $10^{-3}$,
and weight decay $10^{-4}$. Early stopping with patience 3 is based on
source validation accuracy. Experiments use random seed 42 and an
NVIDIA A100 GPU. We use Qwen3-4B  as the LLM planner and deploy it locally through Hugging Face. The planner receives only the structured controller state and selects the next diagnostic action or stop action. The LLM does not access raw IQ samples or modify the trust score or final reliability decision.

For source calibrated acceptance threshold selection, we set the
minimum coverage to $0.5$. For the reliability controller, we use
maximum action and cost budgets of $N_{\max}=4$ and $C_{\max}=8$,
respectively. The ordinal diagnostic costs are set to $1$, $1$, $2$, and $4$ for prototype consistency, signal quality, domain similarity, and counterfactual stability, respectively. For the source calibrated evidence penalties, we use $\lambda_{\rm cf}=0.08$, $\lambda_{\rm proto}=0.08$, $\lambda_{\rm dom}=0.04$, and $\lambda_{\rm qual}=0.03$. The severe evidence activation threshold and penalty scale are set to $\tau_{\rm sev}=0.60$ and $\eta_{\rm sev}=0.50$, respectively.

After training, \textsc{TrustAgent-RFF} evaluates prediction without modifying  device label. All diagnostic
statistics, normalization parameters, trust estimators, and decision
thresholds are derived from source training or validation data only.

\subsection{Evaluation Protocol}

We report same domain and cross domain accuracy to quantify physical
domain shift and evaluate whether \textsc{TrustAgent-RFF} identifies
unreliable predictions, especially high confidence errors. Unless
otherwise specified, high confidence means a maximum softmax probability
of at least $0.9$. The controller uses the more conservative threshold
$\tau_{\rm hc}=0.8$, whereas high confidence evaluation uses
$c\geq0.9$.

The main metrics are high confidence error rate, absolute error rate
reduction, coverage, selective accuracy, selective risk, and area under
the risk coverage curve. Coverage is the fraction of retained
predictions, and lower selective risk and area under the risk coverage
curve indicate better reliability ranking. We vary the acceptance
threshold across coverage levels and backbones. Matched coverage comparisons retain the same number of target
predictions across methods, testing whether \textsc{TrustAgent-RFF}
better ranks correct over incorrect predictions rather than relying on
greater rejection.

For individual diagnostics, we report AUROC and AUPRC for separating
correct and incorrect predictions. We also conduct leave one module out
ablations and compare the controller with fixed diagnostic strategies
in reliability, coverage, and diagnostic cost.
\subsection{Source and Target Domain Accuracy}

Table~\ref{tab:source_target_accuracy} demonstrates substantial but
dataset dependent degradation under unseen physical domains. Across the
30 matched dataset backbone pairs, the mean source to target accuracy
drop is 0.387, with a 95\% confidence interval of [0.295, 0.478].
Target accuracy is lower than source accuracy in 93\% of comparisons,
showing that strong source performance rarely persists after physical
domain shift.

The severity of degradation varies across datasets. ORACLE exhibits the
largest average accuracy drop at approximately 0.751, followed by WiSig,
Bluetooth, and LoRa, with average drops of approximately 0.447, 0.346,
and 0.337, respectively. NetSTAR is more stable, with an average gap of
only 0.052. Moreover, CNNGRU and Transformer achieve slightly higher
target than source accuracy on NetSTAR, indicating that domain shift
depends on the protocol and dataset characteristics.

The backbone ranking also changes between source and target domains.
Transformer achieves the highest source accuracy on ORACLE, LoRa, and
Bluetooth, whereas the strongest target models are PatchLLM, SimCLR,
and SimCLR, respectively. On WiSig, SimCLR performs best in the source
domain, while TCN achieves the highest target accuracy. On NetSTAR, TCN
has the highest source accuracy, whereas Transformer performs best on
the target domain. 


These results show that source validation accuracy is not a reliable
indicator of robustness under unseen physical conditions. They motivate
a deployment time reliability mechanism that evaluates individual
predictions using RFF specific evidence. Accordingly, the main benefit
of \textsc{TrustAgent-RFF} is measured through high confidence error
reduction and selective prediction at matched coverage, rather than
through improvements in full sample classification accuracy.

\subsection{Filtering High Confidence Errors under Domain Shift}
\label{subsec:zeroshot_filtering}

The accuracy degradation in
Table~\ref{tab:source_target_accuracy} does not necessarily make the
classifier uncertain. Incorrect target domain predictions can still
receive high maximum softmax probabilities, and increasing the
confidence threshold often reduces errors mainly by lowering coverage.
We therefore evaluate whether \textsc{TrustAgent-RFF} can filter
incorrect predictions from the same high confidence candidate set.

For a classifier confidence threshold $\tau$, let
$\mathcal H_\tau=\{x:c(x)\geq\tau\}$ denote the high confidence
candidate set. The softmax baseline accepts all predictions in
$\mathcal H_\tau$, whereas \textsc{TrustAgent-RFF} uses the frozen
classifier, source calibration statistics, and  RF diagnostic
evidence to retain a subset
$\mathcal R_\tau\subseteq\mathcal H_\tau$. The predicted transmitter
identity is not modified.

For any retained set $\mathcal R$, we define
$
\mathrm{HCER}(\mathcal R)
=
\frac{1}{|\mathcal R|}
\sum_{x_i\in\mathcal R}
\mathbb I[\hat y_i\neq y_i].
$
$
\Delta_{\rm HCER}
=
\mathrm{HCER}(\mathcal H_\tau)
-
\mathrm{HCER}(\mathcal R_\tau).
$
A positive $\Delta_{\rm HCER}$ indicates that
\textsc{TrustAgent-RFF} lowers the error rate among retained high
confidence predictions.

\begin{table}[t] \centering \caption{ Predictions with base classifier confidence $\geq 0.9$. Retention is the percentage of high confidence predictions accepted by TrustAgent-RFF. } \label{tab:zeroshot_high_confidence} \small \setlength{\tabcolsep}{5pt} \renewcommand{\arraystretch}{1.10} \begin{tabular}{lcccc} \toprule Dataset & Retention & Softmax & TrustAgent & $\Delta_{\rm HCER} $\\ \midrule Bluetooth & 20.3\% & 13.7\% & 3.7\% & 10.0 \\ LoRa & 99.3\% & 51.9\% & 50.4\% & 1.5 \\ NetSTAR LoRa & 98.1\% & 3.0\% & 2.6\% & 0.4 \\ ORACLE & 51.4\% & 83.9\% & 83.2\% & 0.7 \\ WiSig & 28.4\% & 45.4\% & 22.2\% & \textbf{23.2} \\ \midrule Overall & 59.5\% & 39.6\% & 32.4\% & \textbf{7.2} \\ \bottomrule \end{tabular} \end{table}

Table~\ref{tab:zeroshot_high_confidence} shows that
\textsc{TrustAgent-RFF} retains 59.5\% of high confidence predictions
and reduces their overall error rate from 39.6\% to 32.4\%, an absolute
reduction of 7.2 percentage points. The largest reductions occur on
WiSig and Bluetooth. LoRa and NetSTAR LoRa show smaller improvements
because nearly all high confidence predictions are retained, while
ORACLE remains challenging because incorrect predictions dominate the
candidate set and the retained subset.

\subsection{Selective Prediction at Matched Coverage}
\label{subsec:risk_coverage}

A lower error rate caused by rejecting more predictions does not
necessarily imply better reliability ranking. We therefore compare
\textsc{TrustAgent-RFF} with maximum softmax probability at matched
coverage, where the softmax baseline retains the same number of
predictions according to its confidence scores. For a retained set $\mathcal R$, selective risk is
$
\mathrm{Risk}(\mathcal R)
=
\frac{
\sum_{i\in\mathcal R}
\mathbb{I}[\hat y_i\neq y_i]
}{
|\mathcal R|
}.
$
Lower risk at the same coverage indicates a safer retained subset.

\begin{figure}[t]
\centering

\begin{subfigure}[t]{0.485\columnwidth}
    \centering
    \includegraphics[
        width=\linewidth,
        height=3cm
    ]{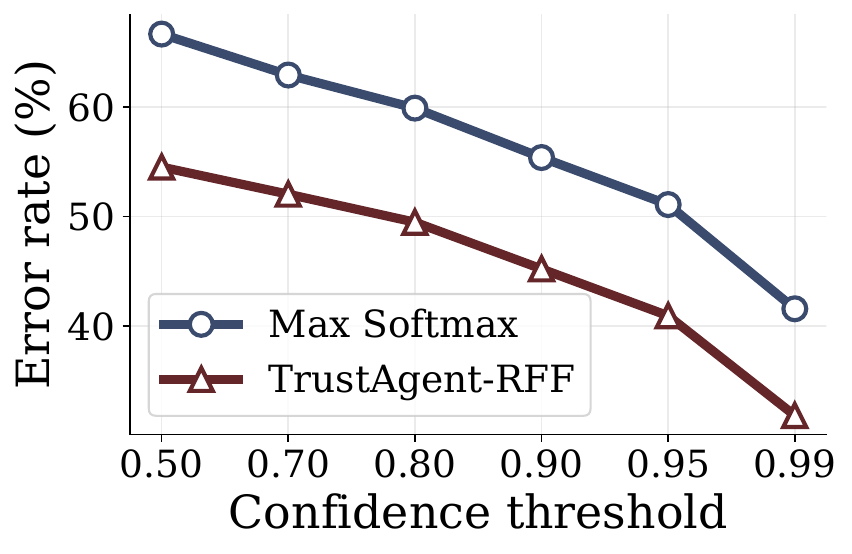}
    \caption{Error rate across confidence thresholds.}
    \label{fig:hcer_thresholds}
\end{subfigure}
\hfill
\begin{subfigure}[t]{0.485\columnwidth}
    \centering
    \includegraphics[
        width=\linewidth,
        height=3cm
    ]{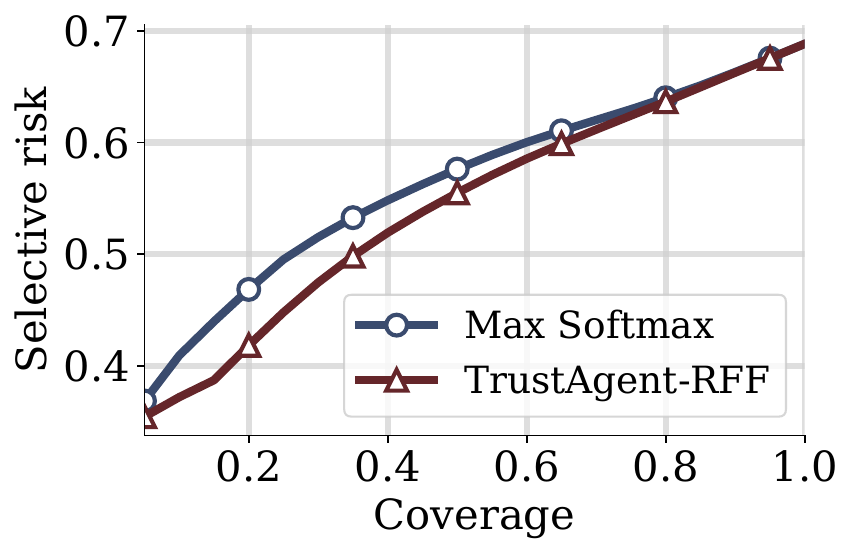}
    \caption{Selective risk at matched coverage.}
    \label{fig:risk_coverage}
\end{subfigure}

\caption{
Reliability evaluation of \textsc{TrustAgent-RFF}.
(a) Retained error rate as the different classifier confidence threshold.
(b) Selective risk for max softmax  and the
\textsc{TrustAgent-RFF} trust score at matched coverage.
Lower values are better.
}
\label{fig:reliability_results}
\end{figure}

Figure~\ref{fig:reliability_results}(a) shows that the improvement is
not limited to a single confidence threshold. As the confidence
threshold increases, both methods retain fewer predictions, but
\textsc{TrustAgent-RFF} consistently achieves a lower retained error
rate, including at $\tau=0.99$. This result confirms that very high softmax confidence does not
eliminate incorrect predictions under physical domain shift. Figure~\ref{fig:reliability_results}(b) further shows that
\textsc{TrustAgent-RFF} achieves lower selective risk at matched
coverage, especially at low and intermediate coverage. As coverage
increases, the risk curves converge and coincide at full coverage
because \textsc{TrustAgent-RFF} does not modify device predictions.

Overall, \textsc{TrustAgent-RFF} reduces high confidence errors while
maintaining lower selective risk, indicating more
effective reliability ranking than softmax confidence.

\subsection{Consistency across Backbones}
\label{subsec:backbone_consistency}

To examine whether the reliability gains depend on a particular classifier architecture, we evaluate \textsc{TrustAgent-RFF} across multiple backbones.

\begin{figure}[t]
\centering
\includegraphics[
  width=0.98\columnwidth,
  trim={0mm 0mm 0mm 0mm},
  clip
]{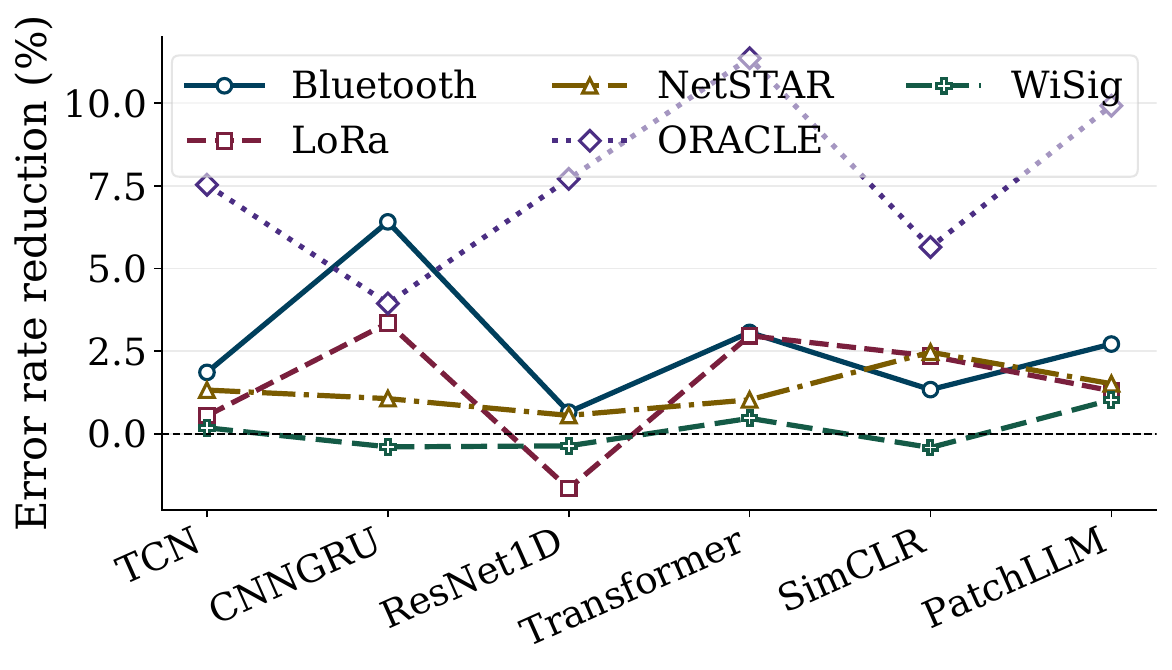}
\caption{
Absolute reduction in high confidence error rate across backbones and
datasets. Positive values indicate improvement, while negative values
indicate degradation.
}
\label{fig:backbone_generalization}
\end{figure}

Figure~\ref{fig:backbone_generalization} evaluates CNN-GRU, TCN,
ResNet1D, Transformer, SimCLR, and PatchLLM across recurrent,
convolutional, attention based, self supervised, and patch based
architectures. \textsc{TrustAgent-RFF} achieves positive error rate
reduction for most dataset and backbone combinations, indicating that
its benefit is not limited to a specific classifier family. The improvement is strongest on ORACLE, where all six backbones benefit,
with Transformer showing the largest reduction. Bluetooth also shows
consistent gains, while NetSTAR exhibits smaller improvements. LoRa
shows degradation for ResNet1D, and WiSig is less consistent, with
clearer gains mainly for Transformer and PatchLLM.

Overall, \textsc{TrustAgent-RFF} generalizes across classifier
architectures, while the improvement depends more strongly on the
dataset and physical domain shift. Larger gains occur when prediction
errors produce detectable inconsistencies in the RF diagnostic evidence.
\subsection{Runtime Overhead}
\label{subsec:runtime}

\begin{table}[t]
\centering

\begin{minipage}[t]{0.48\columnwidth}
\vspace{0pt}
\centering

\begin{minipage}[t][2.4\baselineskip][t]{\linewidth}
\captionof{table}{Average inference latency per sample.}
\label{tab:runtime_overhead}
\end{minipage}

\vspace{2pt}
\small
\setlength{\tabcolsep}{5pt}
\renewcommand{\arraystretch}{1.00}
\begin{tabular}{lcc}
\toprule
Method & Latency & Ratio \\
\midrule
Base
& 1.84
& 1.0$\times$ \\

Ours
& 29.27
& 15.9$\times$ \\
\bottomrule
\end{tabular}

\end{minipage}
\hfill
\begin{minipage}[t]{0.48\columnwidth}
\vspace{0pt}
\centering

\begin{minipage}[t][2.4\baselineskip][t]{\linewidth}
\captionof{table}{LLM Diagnostic cost comparison.}
\label{tab:diagnostic_cost}
\end{minipage}

\vspace{2pt}
\small
\setlength{\tabcolsep}{5pt}
\renewcommand{\arraystretch}{1.00}
\begin{tabular}{lcc}
\toprule
Policy & Actions & Cost \\
\midrule
Sequential
& 4.000
& 8.000 \\

Ours
& 2.153
& 3.658 \\
\bottomrule
\end{tabular}

\end{minipage}

\end{table}
As shown in Table~\ref{tab:runtime_overhead},  \textsc{TrustAgent-RFF} diagnostics increases the average inference
latency from 1.84~ms to 29.27~ms per sample. The main overhead comes from counterfactual stability, which requires multiple additional
classifier evaluations. This result motivates adaptive diagnostic selection rather than executing every module for every prediction.

Table~\ref{tab:diagnostic_cost} compares sequential action costs between All modules and \textsc{TrustAgent-RFF}. The all modules policy executes all
four diagnostics and incurs a fixed assigned cost of 8 per sample. In
contrast, Our controller executes an average of 2.153 actions and incurs an average assigned cost of 3.658. This reduces the
number of executed actions by 46.2\% and the assigned diagnostic cost by
54.3\%. The result shows that many predictions can reach a terminal decision without acquiring all diagnostic evidence, improving
efficiency while preserving the reliability benefits of the full
framework.
\subsection{Ablation Study}
We conduct four ablation studies. First, we compare the full
\textsc{TrustAgent-RFF} framework with a modules only variant to evaluate
the contribution of the complete trust assessment pipeline. Second, we
remove each diagnostic module individually to examine whether the four
diagnostics provide complementary reliability evidence. Third, we
compare diagnostic selection with sequential execution to
evaluate whether the controller reduces diagnostic cost.

\begin{figure}[t]
    \centering
    \begin{subfigure}[t]{0.49\columnwidth}
        \centering
        \includegraphics[width=\linewidth]{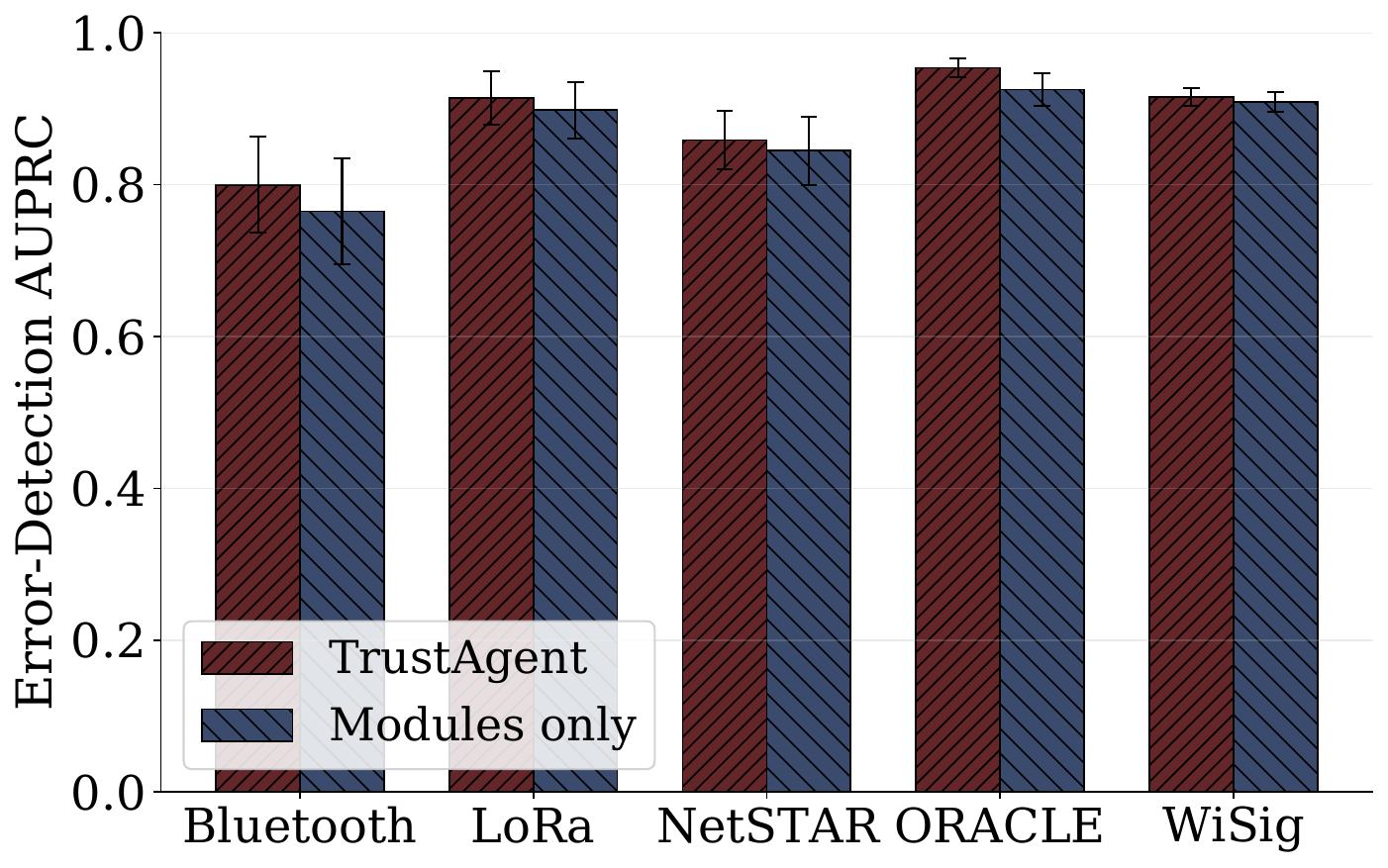}
        \caption{Error detection AUPRC.}
        \label{fig:trustagent_auprc}
    \end{subfigure}
    \hfill
    \begin{subfigure}[t]{0.49\columnwidth}
        \centering
        \includegraphics[width=\linewidth]{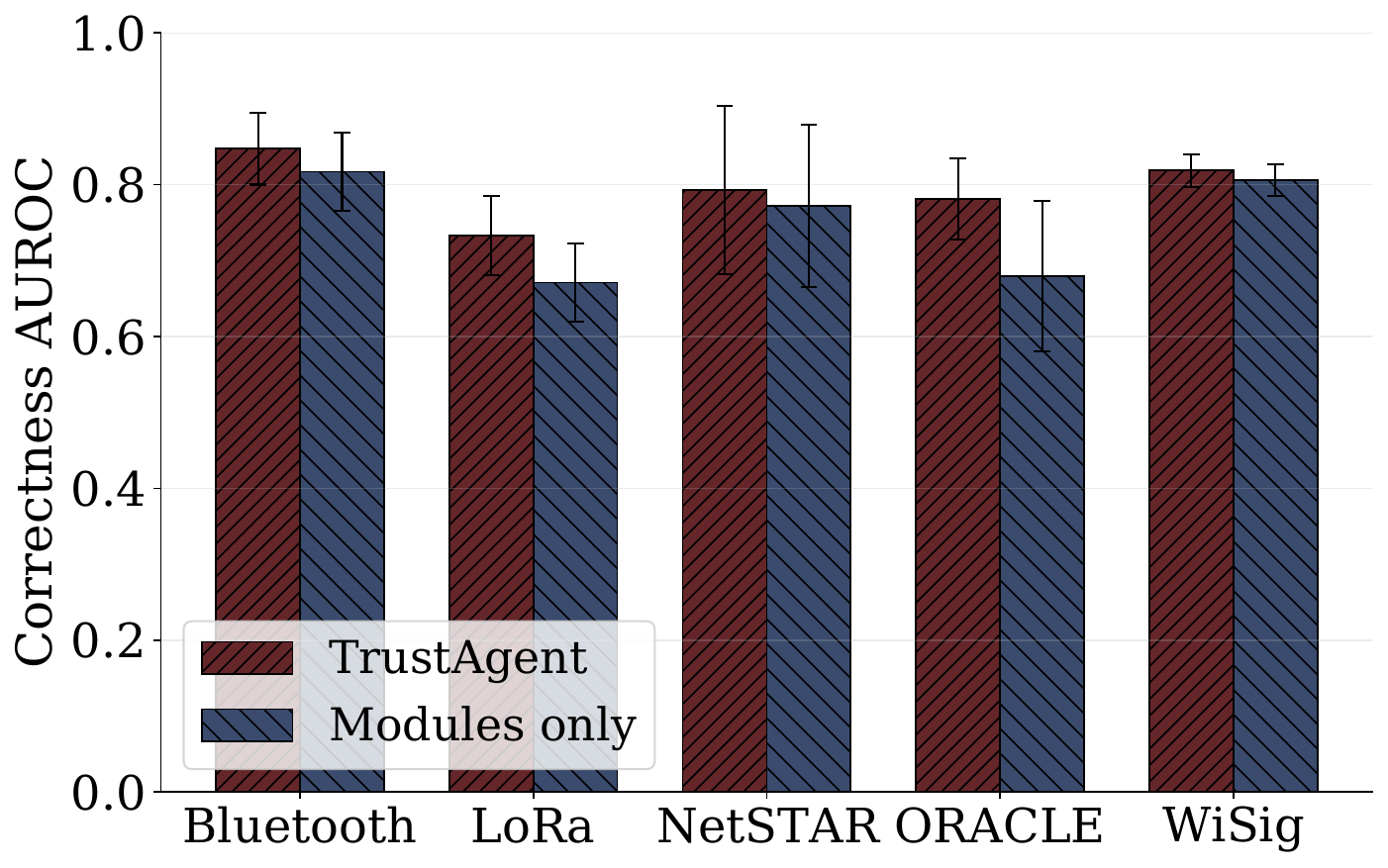}
        \caption{Correctness AUROC.}
        \label{fig:trustagent_auroc}
    \end{subfigure}
    \caption{
\textsc{TrustAgent-RFF} versus modules only:
(a) correctness AUROC and (b) error detection AUPRC.
Bars show backbone means and standard deviations.
}
    \label{fig:trustagent_ablation}
\end{figure}

Figure~\ref{fig:trustagent_ablation} shows that the full framework
achieves higher error detection AUPRC and correctness AUROC across all
five datasets. The improvement is most visible on ORACLE and LoRa,
while smaller gains are observed on Bluetooth, NetSTAR, and WiSig.
These results indicate that the complete trust assessment pipeline
provides a more effective ranking of reliable and unreliable
predictions than using the diagnostic modules alone. We next evaluate the contribution of each diagnostic by removing one
module at a time and comparing selective error rate at matched coverage.

\begin{table}[t]
\centering

\caption{First-action AURC of LLM planners averaged across six
backbones. Lower is better.}
\label{tab:llm_first_action_aurc}
\small
\setlength{\tabcolsep}{4pt}
\renewcommand{\arraystretch}{1.00}

\resizebox{\columnwidth}{!}{
\begin{tabular}{lccccc}
\toprule
Planner & Bluetooth & LoRa & NetSTAR & ORACLE & WiSig \\
\midrule
SmolLM3-3B
& 0.178
& 0.737
& 0.390
& 0.715
& 0.666 \\

Llama-3.2-3B
& 0.182
& 0.742
& 0.388
& 0.720
& 0.672 \\

Qwen3-4B
& 0.176
& 0.737
& 0.391
& 0.705
& 0.669 \\
\bottomrule
\end{tabular}
}

\end{table}

Table~\ref{tab:llm_first_action_aurc} compares the first action
performance of different LLM planners across datasets. No single
planner consistently dominates all datasets. Qwen3 achieves the lowest
AURC on Bluetooth and ORACLE and ties with SmolLM3 on LoRa. SmolLM3
performs best on WiSig, while Llama achieves the lowest AURC on
NetSTAR. The differences among the three planners are generally small,
indicating that \textsc{TrustAgent-RFF} is relatively robust to the
choice of LLM planner.

\begin{table}[t]
\centering
\caption{Module ablation in selective error rate at matched coverage.
Lower is better.}
\label{tab:module_ablation_error}
\resizebox{\columnwidth}{!}{
\begin{tabular}{l|ccccc}
\toprule
& \multicolumn{5}{c}{Selective Error Rate (\%) $\downarrow$} \\
\cmidrule(lr){2-6}
Variant
& Bluetooth
& LoRa
& NetSTAR
& ORACLE
& WiSig \\
\midrule
TrustAgent
& \textbf{9.29}
& \textbf{74.68}
& \textbf{42.05}
& \textbf{82.06}
& \textbf{18.93} \\

w/o CF
& 11.23
& 75.26
& 42.18
& 84.03
& 26.01 \\

w/o Proto
& 14.53
& 74.97
& 42.27
& 83.39
& 19.31 \\

w/o Dom
& 9.67
& 74.95
& 42.09
& 83.17
& 22.42 \\

w/o Qual
& 9.58
& 75.11
& 42.12
& 83.75
& 22.44 \\
\bottomrule
\end{tabular}
}
\end{table}


As shown in Table~\ref{tab:module_ablation_error}, the complete
\textsc{TrustAgent-RFF} framework achieves the lowest selective error
rate on all five datasets. Ablation effects are larger on Bluetooth and
WiSig, where prototype consistency and counterfactual/domain evidence
contribute more strongly, while LoRa and NetSTAR remain more difficult.
Overall, the four diagnostics provide complementary and dataset-dependent
reliability evidence.
\section{Conclusion}

This paper presented \textsc{TrustAgent-RFF}, a post hoc reliability
framework for RF fingerprinting under unseen physical domain shifts.
Operating after a frozen classifier, it combines four complementary RF
diagnostics with a source calibrated trust estimator and uses an LLM
planner to acquire evidence under a computational budget.
Experiments across five datasets and six classifier backbones reveal
substantial cross domain degradation and persistent high confidence
errors. \textsc{TrustAgent-RFF} reduces the overall error rate among
retained high confidence predictions from 39.6\% to 32.4\%, achieves
lower selective risk than softmax confidence at matched coverage, and
reduces the assigned diagnostic cost by 54.3\% relative to sequential
execution.

\bibliographystyle{IEEEtran}
\bibliography{references}
\vspace{12pt}

\end{document}